\documentclass[letterpaper, 10 pt, conference]{ieeeconf}
\IEEEoverridecommandlockouts

\usepackage[usenames,dvipsnames]{xcolor}

\usepackage{amsmath,amssymb,amsfonts}
\usepackage{algorithmic}
\usepackage{graphicx}
\usepackage{textcomp}
\usepackage{multirow}
\usepackage{xcolor}
\usepackage{colortbl}
\definecolor{lightpeach}{rgb}{1.0, 0.87, 0.80} 
\definecolor{asparagus}{rgb}{0.4, 0.69, 0.2}
\definecolor{yellowcalm}{rgb}{0.97, 0.91, 0.56}
\usepackage{leftidx}
\usepackage{color,soul}
\usepackage{caption}
\usepackage{subcaption}
\usepackage{makecell}
\usepackage[export]{adjustbox}
\usepackage[sort,compress]{cite}
\usepackage[pdftex, colorlinks=true, linkcolor=black, citecolor = blue, urlcolor = blue, filecolor=red]{hyperref}
\usepackage{booktabs}
\usepackage[nolist]{acronym}
\usepackage{tikz}
\usepackage[finnish]{babel}
\usepackage{gensymb,balance}

\newacro{slam}[SLAM]{Simultaneous Localization and Mapping}
\newacro{uav}[UAV]{Unmanned Aerial Vehicle}
\acrodefplural{uav}[UAVs]{unmanned aerial vehicles}
\newacro{gns}[GNS]{Global Navigation Satellite}
\newacro{gnss}[GNSS]{Global Navigation Satellite System}
\acrodefplural{gnss}[GNSS]{Global Navigation Satellite Systems}
\newacro{mcl}[MCL]{Monte-Carlo localization}
\newacro{imu}[IMU]{Inertial Measurement Unit}
\newacro{dof}[DOF]{degree-of-freedom}
\acrodefplural{dof}[DOFs]{degrees-of-freedom}
\newacro{ransac}[RANSAC]{random sample consensus}
\newacro{map}[MAP]{maximum a posteriori}
\newacro{mle}[MLE]{maximum likelihood estimation}
\newacro{rms}[RMS]{root-mean-square}
\newacro{vio}[VIO]{visual-inertial odometry}
\newacro{iou}[IoU]{intersection over union}
\newacro{ekf}[EKF]{extended Kalman filter}
\newacro{swap}[SWaP]{size, weight and power}
\newacro{sfm}[SFM]{Structure-from-Motion}
\newacro{sam}[SAM]{Segment Anything Model}
\newacro{vpr}[VPR]{Visual Place Recognition}
\newacro{lwir}[LWIR]{Long-Wavelength Infrared}

\newcommand{\figref}[1]{\hyperref[#1]{Figure~\ref*{#1}}}
\newcommand{\tabref}[1]{\hyperref[#1]{Table~\ref*{#1}}}
\newcommand{\secref}[1]{\hyperref[#1]{Sec.~\ref*{#1}}}
\newcommand{\algoref}[1]{\hyperref[#1]{Alg.~\ref*{#1}}}

\def\ie{\textit{i.e.},}
\def\eg{\textit{e.g.},}

\def\batvik{B\aa{}tvik}

\def\methodabbreviation{SOS-Match}

\def\cameraimage{\ensuremath{\mathcal{I}}}
\def\time{\ensuremath{t}}
\def\segmentationmaskindex{\ensuremath{k}}
\def\segmentationmaskindexmax{\ensuremath{K}}

\newcommand{\segmentationmask}[1]{\ensuremath{I_{#1}}}
\newcommand{\centroidpxcoordinates}[1]{\ensuremath{m_{#1}}}\newcommand{\pxcoordinatecovariance}[1]{\ensuremath{\Sigma_{#1}}}
\def\setofsiftfeatures{\ensuremath{A}}

\def\sizedescriptor{\ensuremath{h}}
\def\eigenvalue{\ensuremath{\lambda}}

\def\meanofviopixelmotion{\ensuremath{\mu_p}}
\def\stdofviopixelmotion{\ensuremath{\sigma_p}}
\def\camerapose{\ensuremath{T}}
\def\indexsymboli{\ensuremath{i}}
\def\indexsymbolj{\ensuremath{j}}

\def\specialeuclideangroup3{\ensuremath{SE(3)}}

\def\relativesize{\ensuremath{r}}
\def\scoringfunction{\ensuremath{q}}
\def\sizescoringfunction{\ensuremath{q_s}}
\def\featurescoringfunction{\ensuremath{q_f}}
\newcommand{\parameter}[1]{\theta_{#1}}
\def\landmarkposition{\ensuremath{l}}
\def\landmarkindex{\ensuremath{n}}
\def\landmarkmaxindex{\ensuremath{N}}
\def\windowlength{\ensuremath{\parameter{WL}}}
\def\stride{\ensuremath{\parameter{SL}}}
\newcommand{\vehiclemap}[1]{\ensuremath{\mathcal{M}_{v,#1}}}
\def\clipperepsilon{\ensuremath{\varepsilon}}
\newcommand{\mapstrideoffset}[1]{\ensuremath{a_{#1}}}
\def\clipperreturnedset{\ensuremath{S}}

\title{\methodabbreviation: Segmentation for Open-Set Robust Correspondence Search and Robot Localization in Unstructured Environments}

\author{Annika Thomas$^{1*}$, Jouko Kinnari$^{2*}$, Parker C.\ Lusk$^{1}$, Kota Kondo$^{1}$, and Jonathan P.\ How$^{1}$
\thanks{$^*$ Equal Contribution}
\thanks{$^{1}$A. Thomas, K. Kondo and J. How are with the Department of Aeronautics and Astronautics, Massachusetts Institute of Technology.
	    {\texttt{\{annikat, plusk, kkondo, jhow\}@mit.edu.}}}
\thanks{$^{2}$J. Kinnari is with Saab Finland Oy,
Salomonkatu 17B, 00100 Helsinki, Finland
{\tt\small jouko.kinnari@saabgroup.com}}%
}

\begin{document}

\maketitle


\begin{abstract}

We present \methodabbreviation\/, a novel framework for detecting and matching objects in unstructured environments. Our system consists of 1) a front-end mapping pipeline using a zero-shot segmentation model to extract object masks from images and track them across frames and 2) a frame alignment pipeline that uses the geometric consistency of object relationships to efficiently localize across a variety of conditions. We evaluate \methodabbreviation\ on the \batvik{} seasonal dataset which includes drone flights collected over a coastal plot of southern Finland during different seasons and lighting conditions. Results show that our approach is more robust to changes in lighting and appearance than classical image feature-based approaches or global descriptor methods, and it provides more viewpoint invariance than learning-based feature detection and description approaches. \methodabbreviation\ localizes within a reference map up to 46x faster than other feature-based approaches and has a map size less than 0.5\% the size of the most compact other maps. \methodabbreviation\ is a promising new approach for landmark detection and correspondence search in unstructured environments that is robust to changes in lighting and appearance and is more computationally efficient than other approaches, suggesting that the geometric arrangement of segments is a valuable localization cue in unstructured environments. We release our datasets at \url{https://acl.mit.edu/SOS-Match/}.

\end{abstract}

\acresetall

\section{Introduction}
\label{sec:introduction}

The capability of a robot to localize itself with respect to an environment is a fundamental requirement in mobile robotics. Various approaches exist for achieving this, including infrastructure-based methods, map-based methods, and \ac{slam}. 

Infrastructure-based methods such as \ac{gnss} directly provide estimates of location in a known coordinate system but are subject to interference by malicious actors \cite{morales2020coping} and limited in availability (\eg{} only work outdoors). Map-based methods such as \cite{kinnari2023lsvl} allow localization but only in cases where a global map can be acquired of the environment prior to operation. \ac{slam}-based approaches do not depend on the availability of localization infrastructure or a pre-acquired map of the operating environment, and are able to provide a notion of pose with respect to a robot's initial starting position and orientation \cite{cadena2016past}. In multi-agent \ac{slam} cases such as  \cite{tian2022kimeramulti}, there is an additional need to find the alignment between the reference frames of different agents operating in the same environment.

\begin{figure}[t!]
    \centering
    \includegraphics[width=0.48\textwidth]{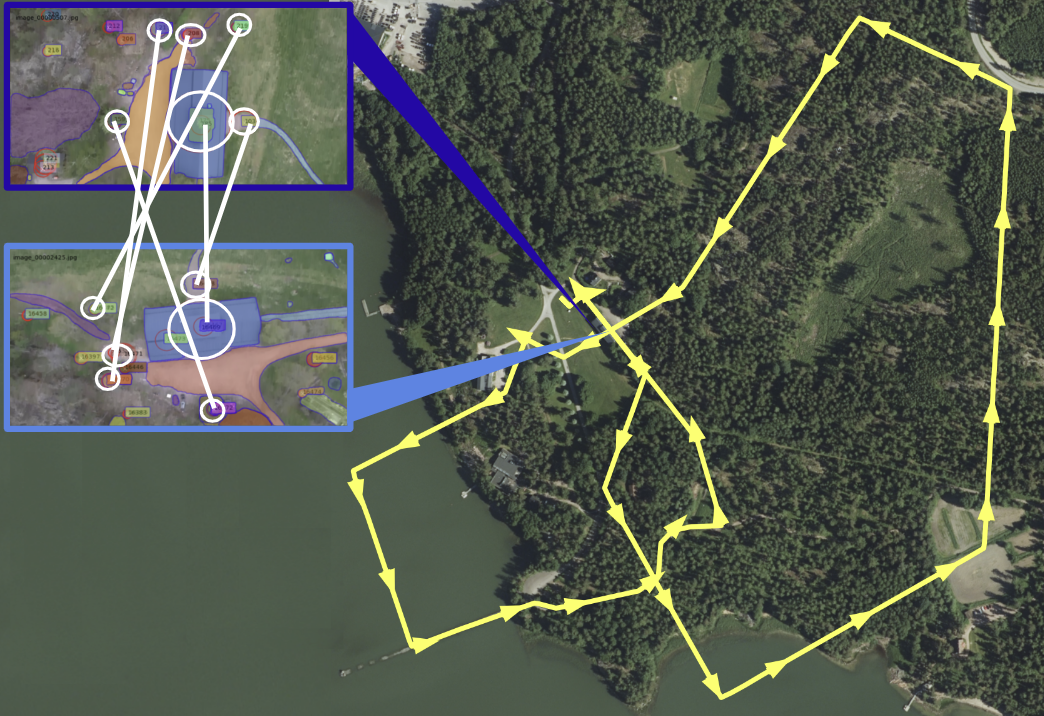}
    \caption{\methodabbreviation\ tracks object masks produced with no pre-training or fine-tuning across sequential posed camera frames to build sparse object-based maps. It robustly associates object masks using their geometric relationship with each other, enabling correspondence detections between traverses over highly ambiguous natural terrains.}
    \label{fig:fig1}
\end{figure}

A fundamental question to address in \ac{slam}, as well as map-based localization approaches, is how to relate the current \emph{environment measurements}, \ie{} sensor inputs in the vicinity of the robot, efficiently and accurately to a reference map. We propose four requirements of robust correspondence search in the localization problem in unstructured environments. To resolve the ambiguities in the correspondence search of current observations and past observations, or across observations by different robots, the description should (1) provide high precision and recall. The method of description should (2) enable operation in an environment which lacks prominent landmarks, operating in a zero-shot approach \ie{} without requiring significant engineering effort if the application domain changes. To allow operation over extended periods, it should (3) be robust to the variation in the appearance of the environment \eg{} due to appearance change over seasonal time in the year. The description of the environment should (4) be modest in terms of memory use, computation time and communication bandwidth requirement for multi-agent scenarios.

To meet these requirements, we present \methodabbreviation{}, an open-set mapping and correspondence search pipeline that makes no prior assumptions about the content of the environment to extract and map objects from visually ambiguous unstructured settings, and uses only the geometric structure of the environment as cue for localization. Utilizing the \ac{sam} \cite{kirillov2023segment} for front-end segment detection, our pipeline tracks detected segments across frames and prunes spurious detections to construct a map of consistent object masks, without requiring additional training per usage environment. We use a robust graph-theoretic data association method \cite{9561069} to associate object locations within object maps, leveraging the geometric arrangement of landmarks and their relative position as cues for localization. Since many localization algorithms are expensive to run on platforms with limited computing resources, we formulate a windowed correspondence search that can trade off accuracy for computational cost. This is an especially suitable approach for drone localization over unstructured terrain, as we demonstrate through experiments with localization and loop closure detection in drone flights over time.

In summary, the contributions of this work include:
\begin{itemize}
    \item A front end capable of reconstructing vehicle maps made of segmented object masks that are less than $0.5$\% the size of other benchmark maps, relying on no prior assumptions of the operating environment.
    \item A method for relating vehicle maps using a geometric correspondence search with a windowed approach that localizes up to $46$x faster than feature-based data association approaches.
    \item \methodabbreviation\ achieves higher recall compared to classical and learned feature-based methods and a state-of-the-art visual place recognition approach evaluated on real-world flights across varied seasonal and illumination conditions, and provides increased robustness to viewpoint variation in comparison to learned feature-based methods.
    \item We release the \batvik{} seasonal dataset containing long traverses with an \ac{uav} across diverse lighting conditions and seasonal appearance change to promote novel contributions towards localization in unstructured environments.
\end{itemize}

\section{Related work}

Several approaches have been proposed for correspondence search in \ac{slam} and map-based localization. We discuss these approaches by considering various environment representations, segmentation-based approaches, and review deep learning in visual navigation. Additionally, we consider challenges from operation in unstructured environments.

\subsection{Environment Representation}

Descriptive and efficient environment representation lays the framework for robust localization.  In visual \ac{slam}, common environment representations include feature-based or object-based approaches. In feature-based \ac{slam}, the environment is described by consistently detectable features within a series of images like ORB \cite{rublee2011orb}, SIFT \cite{lowe2004distinctive}, SURF \cite{bay2006surf} or learned features \cite{sarlin2020superglue,detone2018superpoint}. Several \ac{slam} systems utilize these features for mapping, as shown in \cite{abaspurkazerouni2022survey}. While feature-based \ac{slam} is widely used, this approach poses a significant data handling challenge due to the substantial data volume it entails. In object-based \ac{slam} methods, object detectors like YOLOv3 \cite{redmon2018yolov3} can be used to extract a set vocabulary of objects in urban environments. Using object-based methods coupled with semantic labels can be advantageous in gathering contextual information about the environment while maintaining a compact map, as demonstrated in \cite{ankenbauer2023global}. 

\subsection{Segmentation}

Segmentation partitions an image into meaningful regions or objects. In computer vision, segmentation is widely used for object detection and classification with classifiers such as YOLOv3 \cite{redmon2018yolov3} or semantic classifications with CLIP \cite{radford2021learning}. Segmentation can be used as a step in methods for path planning and object detection \cite{douillard2011segmentation}. Segment Anything \cite{kirillov2023segment}, uses a trained model to perform segmentation in any environment without assumptions or fine tuning and a modified version \cite{zhao2023fast} has recently been used for coordinate frame alignment in multi-agent trajectory deconfliction \cite{kondo2023puma}.

\subsection{Deep Learning in Visual Navigation}

\begin{figure*}[t]
\centering
\includegraphics[width=2\columnwidth, trim={0 11.5cm 0 0}, clip]{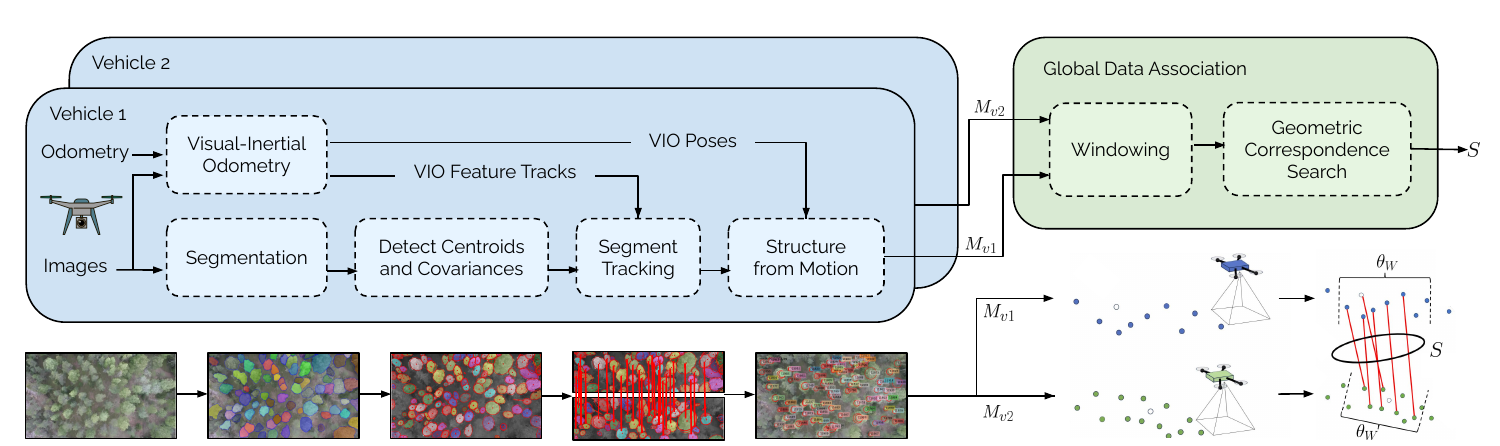}
\caption{\methodabbreviation{} incorporates two novel components. The front end mapping pipeline utilizes the vehicle odometry sensor along with camera images to perform \ac{slam} and generate vehicle maps. The frame alignment pipeline offsets windows and uses our data association algorithm to filter the most likely correspondences.
}
\label{figure_pipeline}
\end{figure*}

Some recent works in deep learning for visual navigation leverage foundational models \cite{bommasani2021opportunities}, which are models trained in a self-supervised way that can accomplish many tasks without fine-tuning or additional training. Deep learning has been used to learn features in methods such as SuperPoint \cite{detone2018superpoint} and D2-Net \cite{dusmanu2019d2}. In visual place recognition, learned global descriptors can facilitate robust scene recognition across viewpoints and differing illuminations. AnyLoc \cite{keetha2023anyloc} is a technique for visual place recognition that uses DINOv2 \cite{oquab2023dinov2}, a self-supervised vision transformer model in combination with unsupervised feature aggregation using VLAD \cite{arandjelovic2013all}, which surpasses other \ac{vpr} approaches in open-set environments, but degrades in cases where contents are similar across frames. 

\subsection{Unstructured Environments}

Unstructured environments are a challenging setting for many visual \ac{slam} systems that assume urban-centric information such as the presence of lane markings or buildings. Some approaches address road roughness and limited distinctive features in these settings by integrating a range of sensor modalities and strategies like using wheel odometry with visual tracking \cite{grimes2009efficient}, integrating topological maps \cite{ort2018autonomous}, and utilizing lidar point clouds \cite{ren2021lidar}. These methods exhibit limitations tied to external hardware requirements, point cloud size constraints, susceptibility to structural changes, and reliance on the assumption of well-defined off-road trails.

\subsection{Placement of This Work}

\methodabbreviation\ utilizes a pre-trained foundation model for segmentation to construct object-based maps without any prior assumptions about the environment such as the presence of objects of specific classes. The generality of this framework allows it to localize successfully in unstructured environments with illumination and structural changes while keeping map sizes compact enough to be shared between multiple agents.

\section{Method}

Our method consists of two main parts; mapping and frame alignment. A block diagram of our method is illustrated in \figref{figure_pipeline} where two vehicles generate object-based maps then perform global data association.

\subsection{Mapping}

Our mapping approach consists of running camera images through a pre-trained image segmentation model such as \cite{kirillov2023segment} or \cite{zhao2023fast}, identifying tracks (\ie{} finding correspondence between object masks across a sequence of images), and reconstructing the positions of the centroids of the segmented areas with a \ac{sfm}-style approach using camera poses estimated by \ac{vio}.

We perform segment detection only after movement of $\parameter{T}$ meters after most recent keyframe, estimated with \ac{vio}. This enables the user of our algorithm the ability to adjust the performance of the system to match the computational resources available on the robot.

Given an image $\cameraimage(\time)$, acquired at time \time, we use the segmentation model to extract binary object masks $\segmentationmask{\segmentationmaskindex}(\time)$, indexed by $\segmentationmaskindex \in \{0, 1, \ldots \segmentationmaskindexmax(\time)\}$. For each $\segmentationmask{\segmentationmaskindex}(\time)$ larger than 2 pixels, we compute the centroid $\centroidpxcoordinates{\segmentationmaskindex}(\time)$ and covariance $\pxcoordinatecovariance{\segmentationmaskindex}(\time)$ of the pixel coordinates of each mask. Further, we extract SIFT features~\cite{Lowe2004} from the region of the mask and store them as set $\setofsiftfeatures_{\segmentationmaskindex}(\time)$. We emphasize that SIFT features are used solely for inter-frame tracking, and we don't keep a record of them in the map. We also compute a size descriptor $\sizedescriptor_{\segmentationmaskindex}(\time) = \sqrt{\eigenvalue_{\segmentationmaskindex}(\time)}$, where $\eigenvalue_{\segmentationmaskindex}(\time)$ is the largest eigenvalue of $\pxcoordinatecovariance{\segmentationmaskindex}(\time)$.

For inter-frame tracking, we assume a generic \ac{vio} implementation is available for tracking camera poses $\camerapose(\time) \in \specialeuclideangroup3$ as well as for tracking the movement of visual feature points between images $\cameraimage(\time_\indexsymboli)$ and $\cameraimage(\time_\indexsymbolj)$. We compute the amount of movement of visual feature points tracked by \ac{vio} in pixels and compute mean \meanofviopixelmotion{} and standard deviation \stdofviopixelmotion{} of the movement.

We evaluate putative correspondences for each track whose latest observation is at most $\parameter{\time}$ keyframes old; this provides some robustness against intermittent temporal inconsistencies in detection of segments by \ac{sam}.

We associate the detections of segments across consecutive image frames using three techniques. First, we require that an epipolar constraint of the segment centroids is satisfied. Second, we require that the apparent shift in pixel coordinates of the segment centroid is in correspondence with the movement of features points tracked by a \ac{vio} algorithm. Third, we match segments that are similar in size and appearance.

For the epipolar constraint (see \eg{} \cite{Hartley2004}), we only allow associations with a margin of less than $\parameter{a}$ pixels. The comparison of the apparent shift to \ac{vio} points is based on only allowing movement less than a specified limit $\parameter{v}$: 

\begin{equation}
\frac{\big||\centroidpxcoordinates{\indexsymboli}(\time-1) - \centroidpxcoordinates{\indexsymbolj}(\time)|-\meanofviopixelmotion\big|}{\stdofviopixelmotion}  < \parameter{v}.
\end{equation}

After excluding infeasible matches based on the epipolar constraint and the requirement of similar movement as \ac{vio} detections, there may still exist more than one possible association between segments observed in latest keyframe to tracks. To this end, we compute the similarity of segment association hypotheses based on their appearance and size. For comparing appearance, we define the feature scoring function $\featurescoringfunction$ as the fraction of SIFT features in sets $\setofsiftfeatures_{\indexsymboli}(\time-1)$ and $\setofsiftfeatures_{\indexsymbolj}(\time)$ that are not eliminated by Lowe's ratio test \cite{Lowe2004}. For comparing sizes, we use a scoring function $\sizescoringfunction(\sizedescriptor_\indexsymboli,\sizedescriptor_\indexsymbolj)$ to weigh each putative association of areas with size descriptors $\sizedescriptor_\indexsymboli$ and $\sizedescriptor_\indexsymbolj$:
\begin{equation}
\label{eq:tracking_size_scoring_function}
\sizescoringfunction(\sizedescriptor_\indexsymboli,\sizedescriptor_\indexsymbolj) =
\begin{cases}
1 + \cos(\frac{\pi}{\parameter{\sizedescriptor}}\relativesize(\sizedescriptor_\indexsymboli,\sizedescriptor_\indexsymbolj)) & \text{if~} \relativesize(\sizedescriptor_\indexsymboli,\sizedescriptor_\indexsymbolj)) < \parameter{\sizedescriptor}, \\
0 & \text{otherwise}
\end{cases}
\end{equation}
where we measure relative size difference of masks $\indexsymboli$ and $\indexsymbolj$ using
\begin{equation}\label{eq:relativesizedifference}
\relativesize(\sizedescriptor_\indexsymboli,\sizedescriptor_\indexsymbolj) = \frac{2|\sizedescriptor_\indexsymboli - \sizedescriptor_\indexsymbolj|}{\sizedescriptor_\indexsymboli + \sizedescriptor_\indexsymbolj}.
\end{equation}
Finally, we compute a similarity score
\begin{equation}
    \scoringfunction = \sqrt{\sizescoringfunction \featurescoringfunction}
    \label{eq:scoringfunction}
\end{equation} as the geometric mean of the size scoring function $\sizescoringfunction$ and the feature scoring function $\featurescoringfunction$. To provide an unambiguous mapping between latest object masks and history of masks, we use an implementation of the Hungarian algorithm \cite{kuhn1955hungarian} using weights from \eqref{eq:scoringfunction}. We initialize new tracks for observations that cannot be matched to previous tracks.

Finally, for tracks with more than $\parameter{\landmarkindex}$ observations, we build a small \ac{sfm}-style factor graph \cite{dellaert2017factor} for each track separately. We specify poses based on odometry and projection factors from the centroid pixel coordinates of segments, and use GTSAM \cite{dellaert2012factor} for finding a minimal cost solution to the factor graph. We record the mean positions of each segment, indexed by $\landmarkindex$ in frame of robot $\indexsymboli$, in odometry frame, $\landmarkposition_{\indexsymboli,\landmarkindex}$. We discard tracks that do not converge to a solution as they are often a result of tracking errors. Furthermore, to describe the size of the segment in a way that is invariant to the distance at which the segment is observed in each image frame, we compute a size descriptor $\sizedescriptor_{S,\landmarkindex}$ scaled approximately to meters, based on the observed pixel size descriptors and distance from the camera to the estimated segment position.

The end result of the mapping pipeline is a vehicle map $\vehiclemap{\indexsymboli}$, which contains estimated positions of objects corresponding to segmented masks, expressed in the odometry frame of robot \indexsymboli{}, and size descriptors for the object masks.

\subsection{Finding correspondences between vehicle maps}

With perfect knowledge of the correspondences between objects, any alignment errors could be mitigated to the level determined by measurement errors of environment measurements. For robot \indexsymboli{} that has observed \landmarkindex{} successfully tracked object masks in $\vehiclemap{\indexsymboli}$, we thus focus on finding the correspondences between objects within its own map $\vehiclemap{\indexsymboli}$  and another map, communicated by a peer or collected at an earlier time instant $\vehiclemap{\indexsymbolj}$.

Assuming no further prior information on the correspondences, the number of possible associations grows quadratically as the number of objects increases, leading to an infeasible search time for any reasonably large map. To utilize the notion that objects spatially close to each other in $\vehiclemap{\indexsymboli}$ should be spatially close to each other in $\vehiclemap{\indexsymbolj}$, we implement a windowed search approach, where we define a window length of \windowlength{} objects, and search for correspondences between the frames, moving forward the window by a stride length of \stride{} objects after each comparison.

We denote $\vehiclemap{\indexsymboli}=\{\landmarkposition_{\indexsymboli,\landmarkindex}\}$ for robot $\indexsymboli$, where $\landmarkindex \in \{0, 1, \ldots, \landmarkmaxindex_\indexsymboli\}$. The windowed search thus attempts to find correspondences between subsets $\vehiclemap{\indexsymboli}[\mapstrideoffset{\indexsymboli} \cdot \stride, ..., \mapstrideoffset{\indexsymboli} \cdot \stride + \windowlength]$ and $\vehiclemap{\indexsymbolj}[\mapstrideoffset{\indexsymbolj} \cdot \stride, ..., \mapstrideoffset{\indexsymbolj} \cdot \stride + \windowlength]$ across all values of $\mapstrideoffset{\indexsymboli}$ and $\mapstrideoffset{\indexsymbolj}$, where the $G[\cdot]$ notation corresponds to taking a subset of $G$ using items with indices $[\cdot]$.

In finding correspondences, we first exclude hypothetical pairs of objects where pairwise difference in distance of the objects in each map is
$\clipperepsilon > \parameter{\clipperepsilon}$ or where size difference of objects is significant, \ie{} $\relativesize(\sizedescriptor_{S,\indexsymboli},\sizedescriptor_{S,\indexsymbolj}) > \parameter{\relativesize}$. We weigh putative associations using \eqref{eq:tracking_size_scoring_function}. We use a robust geometric data association framework \cite{9561069} to approximate a set $\clipperreturnedset$ of associations (object pairs). As a final step, we estimate with \cite{arun1987leastsquares} the relative translation and rotation between $\vehiclemap{\indexsymboli}$ and $\vehiclemap{\indexsymbolj}$, assuming correspondences defined by $\clipperreturnedset$ and discard hypotheses that would result in more than $\parameter{\alpha}$ angular difference in roll or pitch. This is motivated by that odometry frames' roll and tilt can be estimated with an IMU due to excitation from gravity. We use the number of associations returned by the framework, $|\clipperreturnedset|$ as criteria for accepting or rejecting the hypothesis. We accept the hypothesis if $|\clipperreturnedset|>\parameter{\clipperreturnedset}$. By varying threshold $\parameter{\clipperreturnedset}$ for acceptance, we can balance precision and recall of our solution.

\section{Experiments}

\begin{figure}[t!]
    \centering
    \vspace{5pt}
    \includegraphics[trim= {7.5cm 0 6.5cm 1cm}, clip, width=.38\textwidth]{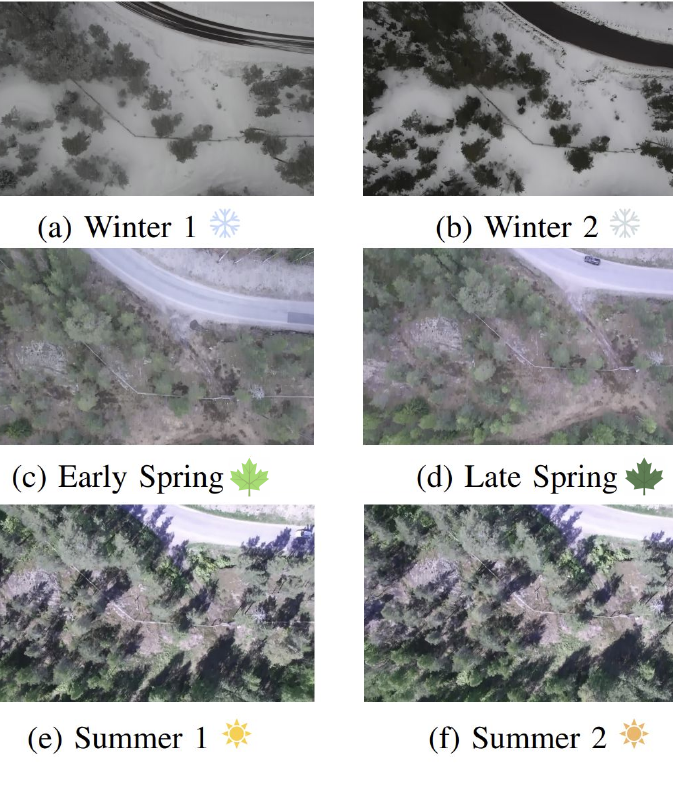}
    \vspace*{-0.1in}
    \caption{Example images from \batvik{} seasonal dataset, including variation in snow coverage, deciduous tree foliage and sharpness of shadows across different seasons.}
    \label{fig:batvikpics}
\end{figure}

We evaluate the performance of \methodabbreviation\ at varying levels of appearance change in the environment, comparing precision, recall, average F1 score, search time and map size with respect to five reference methods. In each experiment, we use a dataset collected for this task.

\subsection{Dataset}

We introduce the \batvik{} seasonal dataset which includes six $3.5$ km drone flights following the same trajectory plan as shown in \figref{fig:fig1} at approximately $100$ m above ground, over many seasonal conditions as illustrated in \figref{fig:batvikpics} and outlined in \tabref{tab:batvik_dataset_flights}. The flights consist of drone images collected with a nadir-pointing camera as well as \ac{imu} measurements, and we record autopilot output along with other telemetry data from an Ardupilot-based \cite{ardupilot} drone flight controller. The flight takes place over an area that contains only a few buildings, and a large part of the trajectory takes place over a forest region, as well as above sea. This dataset represents flight of an \ac{uav} over a terrain that has naturally high ambiguity. 

\subsection{Baseline approaches}

Several visual \ac{slam} approaches \cite{abaspurkazerouni2022survey} use image features such as ORB or SIFT as front end, for detecting and describing feature points, and use \ac{ransac} \cite{fischler1981random} to prune outliers. We implement two baseline methods that detect and describe image features with each approach using keyframes taken every $2$ m of travel of the flight sequence. Next, we use RANSAC to find correspondences that are consistent with respect to the fundamental matrix, and set a limit for the number of required associations to consider the keyframes a match. We extract $500$ SIFT or ORB features, select $20$\% of best matches in terms of descriptors, and use a reprojection threshold of $5.0$ pixels in fundamental matrix filtering. We run \ac{ransac} for a maximum of $2000$ iterations with confidence level $0.995$. 

To compare against state-of-the-art learned detector and descriptor methods, we evaluate against LoFTR \cite{sun2021loftr} with pretrained outdoor weights and the SuperPoint detector \cite{detone2018superpoint} using SuperGlue with pretrained outdoor weights from \cite{sarlin2020superglue} for correspondence search. For each method, we retain only keypoint correspondences with confidence of at least $0.7$, and add all keypoint match values as a metric for the overall match confidence of each image pair. 

\begin{table}[t]
    \centering
    \vspace{5pt}
    \caption{Description of flight trajectories in \batvik{} dataset.}
    \begin{tabular}{p{0.17\linewidth}p{0.02\linewidth}p{0.25\linewidth}p{0.35\linewidth}}
        \toprule
        Name & Icon & Time of flight & Description of appearance \\
        \midrule
        Winter 1 & \includegraphics[trim={8cm 0.5cm 8cm 1cm}, clip, width=0.4cm]{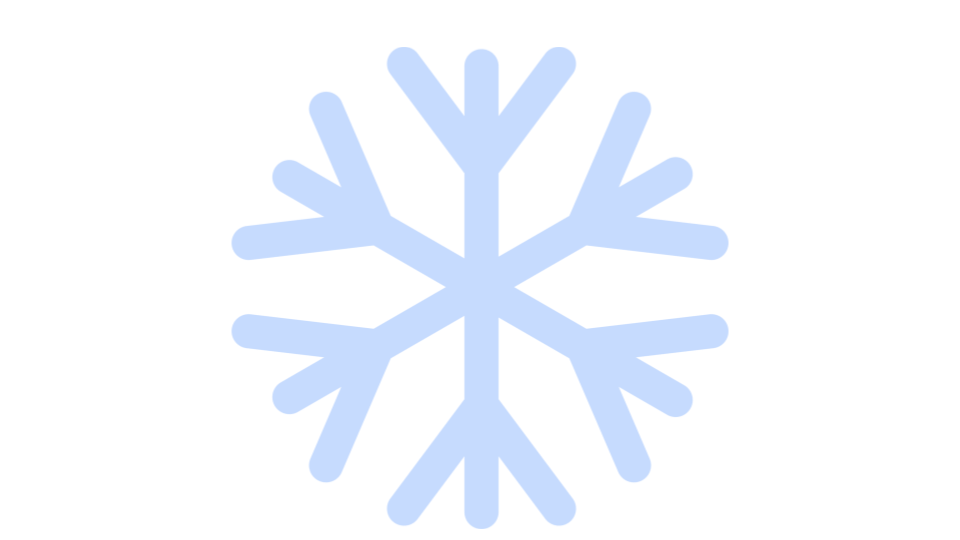} & 2022-03-30 12:51 & Snow coverage \\
        Winter 2 & \includegraphics[trim={8cm 0.5cm 8cm 1cm}, clip, width=0.4cm]{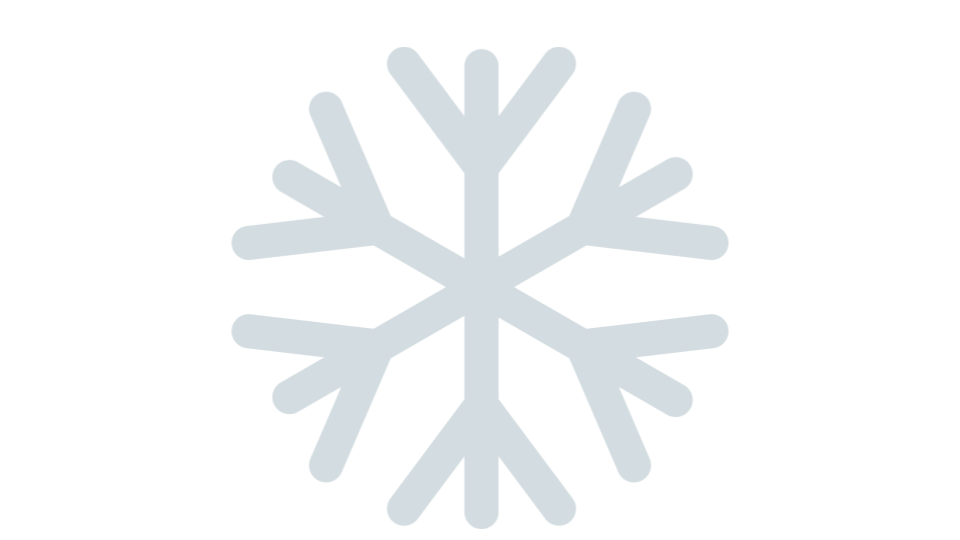} & 2022-03-31 11:39 & Snow coverage \\
        Early Spring & \includegraphics[width=0.4cm]{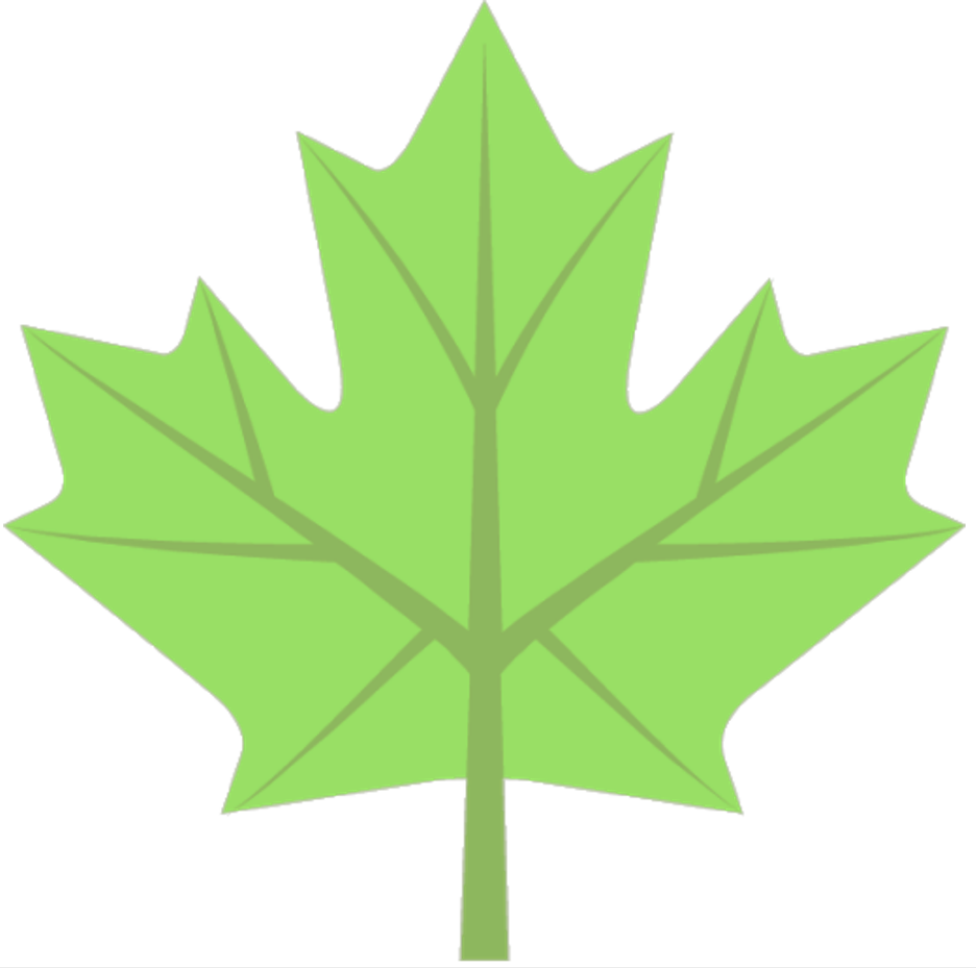} & 2022-05-05 14:10 & Some leaves\\
        Late Spring & \includegraphics[ width=0.4cm]{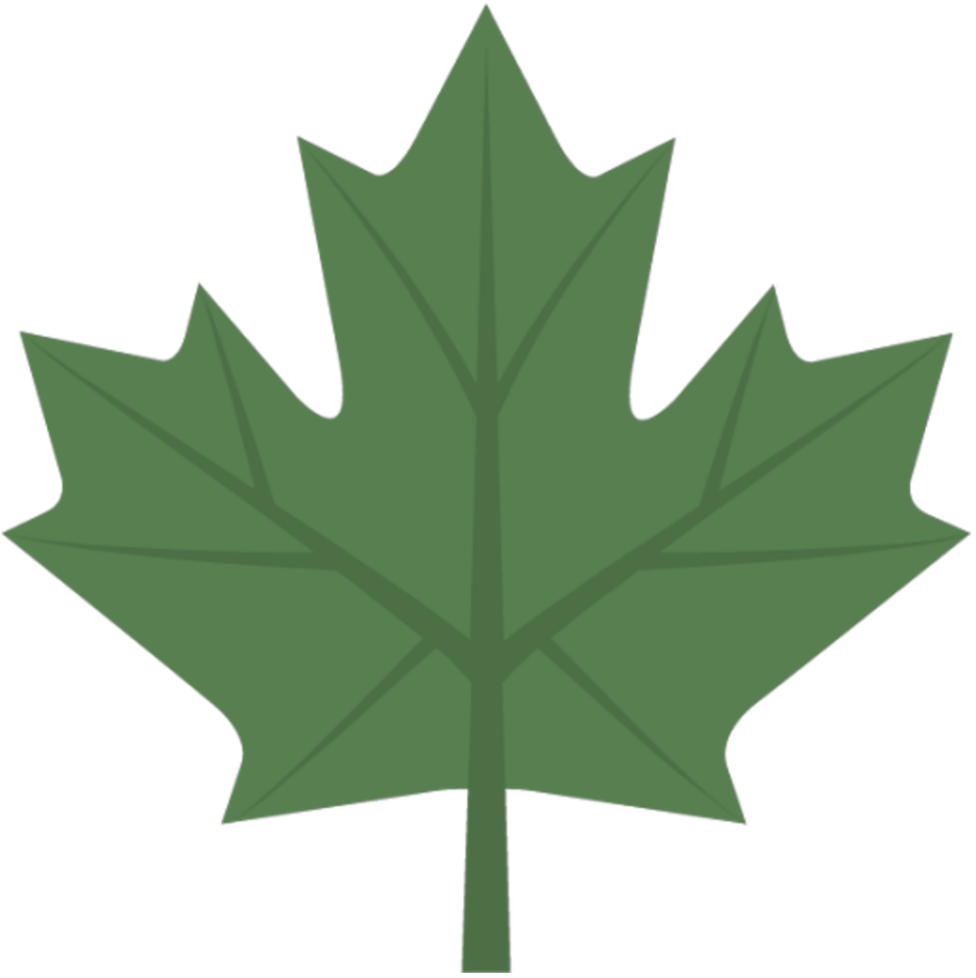} & 2022-05-25 12:33 & Leaves in deciduous plants\\
        Summer 1 & \includegraphics[trim={8cm 0.5cm 8cm 1cm}, clip, width=0.4cm]{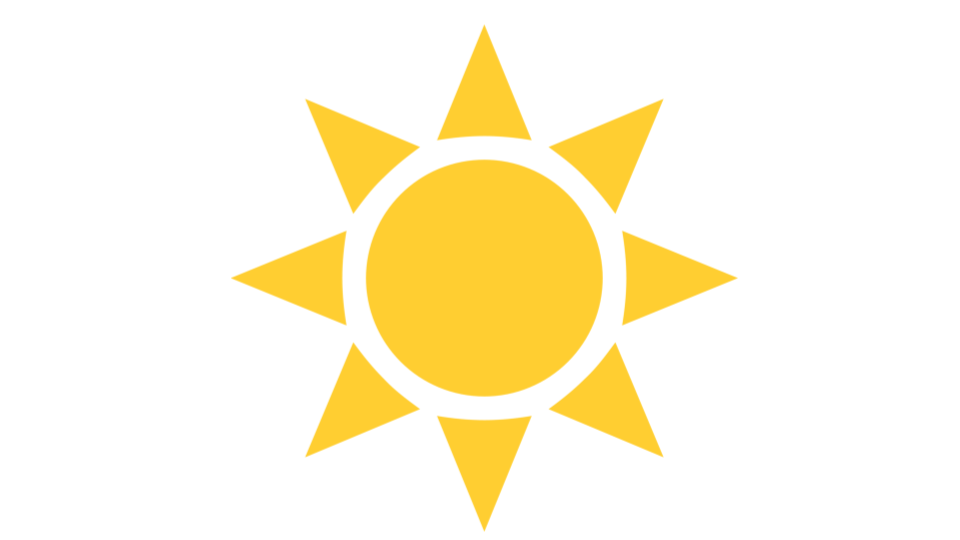} & 2022-06-09 12:05 & Full leaves, hard shadows \\
        Summer 2 & \includegraphics[trim={8cm 0.5cm 8cm 1cm}, clip, width=0.4cm]{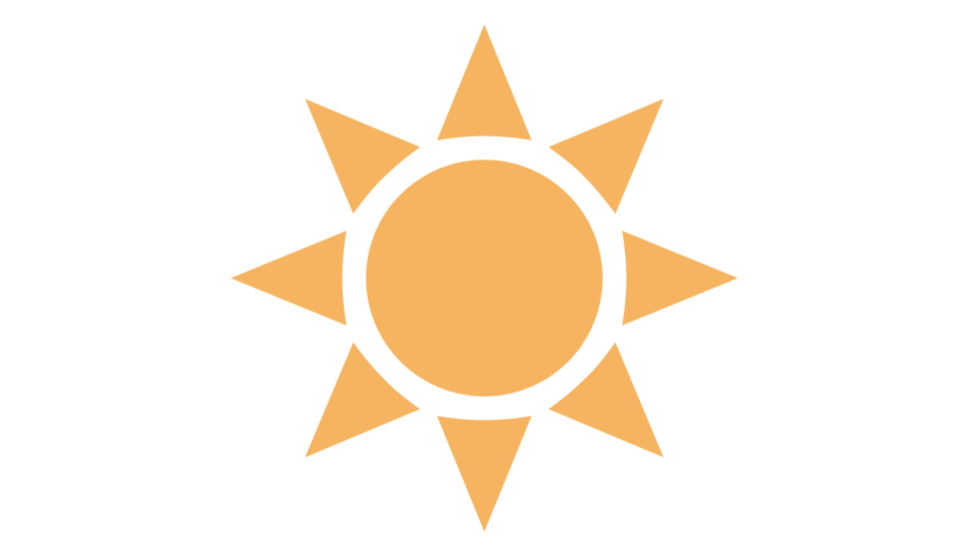} & 2022-06-09 12:28 & Full leaves, hard shadows \\
        \bottomrule
    \end{tabular}
    \label{tab:batvik_dataset_flights}
\end{table}

In addition to feature-based approaches, we compare our method against a modern \ac{vpr} approach that uses global descriptors for images, AnyLoc \cite{keetha2023anyloc}. AnyLoc outperforms universal place recognition pipelines NetVLAD \cite{arandjelovic2016netvlad}, CosPlace \cite{berton2022rethinking} and MixVPR \cite{ali2023mixvpr} in almost every evaluation, making it an appropriate benchmark that authors claim works across very different environmental and lighting conditions.
In our implementation, we define a DINOv2 extractor following AnyLoc's parameters at layer $31$ with facet value and $32$ clusters. We train a VLAD vocabulary of $32$ cluster centers on database images, generate global descriptors for each image in the query set, then compute the cosine similarity of the global descriptors of each image pair in each sequence.

\subsection{Performance measures}

We compare correspondence search results by first computing what region of the ground would be visible from each keyframe camera pose if the ground under the image acquisition position was flat. For this, we use ground truth camera poses recorded from the \ac{ekf} output from a flight controller and a terrain elevation map of the area. By comparing the area of overlap to the area of intersection of each keyframe pairwise, we compute the \ac{iou} of every pair of keyframes. In evaluating recall, we assume that each keyframe pair for which \ac{iou} is more than $0.333$, the matching algorithm should return a match indication. In evaluating precision, we assume that an algorithm may provide a correspondence between frames if the \ac{iou} from ground truth is more than $0.01$; for smaller \acp{iou}, we assume a returned correspondence is a false positive. In mapping, for purposes of evaluation, we use ground truth poses of flight controller \ac{ekf} in \ac{sfm}. We use $3.0$ as pixel measurement noise standard deviation. By varying the runtime as function of window length \windowlength{} and stride \stride{},
we experimentally choose parameters $\windowlength{}=50$ and $\stride{}=10$. For other parameters, we choose 
$\parameter{T} = 2.0$ m,
$\parameter{v} = 4.0$,
$\parameter{\scoringfunction} = 0.2$,
$\parameter{\sizedescriptor} = 0.2$,
$\parameter{\landmarkindex} = 5$,
$\parameter{\alpha}=22.5\degree$,
$\parameter{\clipperepsilon} = 2.0$, and
$\parameter{\relativesize} = 0.2$.

\begin{figure*}[t]

    \includegraphics[trim= {0cm 9.5cm 1.5cm 0cm}, clip, width=1\textwidth]{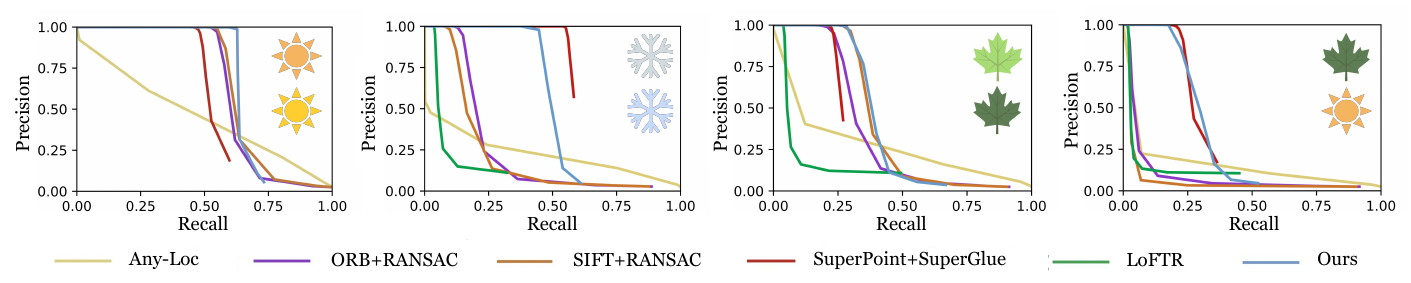}

    \caption{Precision-recall curves with different approaches with increasing visual discrepancy between flights.}
    \label{fig:p-r-curves}
\end{figure*}

We produce precision-recall results by varying the acceptance limit (threshold for number of detected correspondences) for our approach, ORB and SIFT-based methods, the image match confidence threshold for SuperPoint+SuperGlue and LoFTR, and the required level of cosine similarity for AnyLoc.

On an NVIDIA Quadro RTX 3000 with 6 GB VRAM, mean detection and description time is $5.78$ s.
A recent branch of research on the \ac{sam} problem suggests improvements to runtime of the \ac{sam} problem (\eg{} \cite{zhao2023fast}). We forgo detailed discussion of minimizing the front end runtime for our method to focus on the correspondence search runtime characteristics. Our runtime evaluations in all experiments tabulated in \tabref{tab:cross-season} measure time consumed in correspondence search, reflecting time required for localization in real-time settings. The computational time evaluations in \tabref{tab:cross-season} are made with an 2x8 core Intel Xeon 6134 @ $3.2$ GHz cluster computer from which we reserve $16$ GB RAM. For SuperPoint+SuperGlue and LoFTR, which require a GPU for correspondence search, computational time evaluations are made on an NVIDIA RTX 3090.

\subsection{Precision and Recall}

First, we evaluate the performance of \methodabbreviation\ when an agent localizes within a previously collected map from another agent after time has passed. We include sections of the flight in \figref{fig:fig1} that do not involve flying over water, as visual navigation-based systems do not perform well in environments with no distinctive features. Thus, to evaluate the performance of our pipeline over unstructured terrain with a variety of visual features, we consider the flight as a whole in addition to flights from the same viewpoint and different viewpoints. Differentiating into these test cases allows us to evaluate the performance of our method when localizing from the same viewpoint and different viewpoints.

In \figref{fig:p-r-curves}, we show precision and recall of each comparison method localization case after time has passed, increasing visual discrepancy between flights from left to right cases.

\begin{table}[t] 
\scriptsize
\centering
\vspace{5pt}
\caption{Mean search time and map size across flights with increasing visual discrepancy between flights. Best results are highlighted \colorbox{asparagus!50}{first} and \colorbox{yellowcalm!70}{second}, and worst is shown in \color{red}{red}\color{black}{.}}
\vspace*{-0.3em}
\setlength{\tabcolsep}{1.8pt}
\begin{tabular}{c c c c}
	\toprule
	\makecell{Case} & \makecell{Implementation}& \makecell{Mean search time [std] (s)} & \makecell{Map size (Mb)} \\
\toprule
\multirow{6}{*}{
\parbox{1.5cm}{\includegraphics[trim={8cm 0.5cm 8cm 1cm}, clip, width=0.6cm]{images/summer1.png} 
\includegraphics[trim={8cm 0.5cm 8cm 1cm}, clip, width=0.6cm]{images/summer2.png} \\
$\Delta_T$ = 23 min}
}
&  \textbf{Ours}  & \cellcolor{yellowcalm!70}5.34 [0.13] & \cellcolor{asparagus!50}{0.05}\\ 
& ORB+RANSAC & 49.11 [0.25] & \cellcolor{yellowcalm!70}25.43\\ 
& SIFT+RANSAC & {76.82 [0.52]} & {406.95}\\ 
& Any-Loc & \cellcolor{asparagus!50}{0.11 [0.01]} & 310.05 \\ 
& SuperPoint+SuperGlue & {\color{red}166.67 [1.17]} & {\color{red}2593.72} \\
& LoFTR & 133.72 [0.23] & 322.9 \\
\midrule

\multirow{6}{*}{
\parbox{1.5cm}{\includegraphics[trim={8cm 0.5cm 8cm 1cm}, clip, width=0.6cm]{images/winter1.png} 
\includegraphics[trim={8cm 0.5cm 8cm 1cm}, clip, width=0.6cm]{images/winter2.png} \\
$\Delta_T$ = 1 day}
}
&  \textbf{Ours} & \cellcolor{yellowcalm!70}4.35 [0.04] & \cellcolor{asparagus!50}{0.07}\\ 
& ORB+RANSAC &  27.23 [0.48] & \cellcolor{yellowcalm!70}21.94\\ 
& SIFT+RANSAC & {40.28 [0.67]} & 345.89\\ 
& Any-Loc & \cellcolor{asparagus!50}{0.12 [0.01]} & 310.05 \\ 
& SuperPoint+SuperGlue & {\color{red}201.17 [5.88]} & \color{red}2843.62 \\
& LoFTR & 129.66 [0.31] & 454.9 \\
\midrule

\multirow{6}{*}{
\parbox{1.5cm}{\includegraphics[width=0.6cm]{images/sosmatch_earlyspring.pdf}
\includegraphics[width=0.6cm]{images/sosmatch_latespring.pdf} \\
$\Delta_T$ = 20 days}
}
&  \textbf{Ours} & \cellcolor{yellowcalm!70}9.29 [0.12] & \cellcolor{asparagus!50}{0.10}\\ 
& ORB+RANSAC & 38.35 [0.52] & \cellcolor{yellowcalm!70}22.07\\ 
& SIFT+RANSAC & {49.75 [0.67]} & 337.47\\ 
& Any-Loc & \cellcolor{asparagus!50}{0.12 [0.01]} & 310.05 \\ 
& SuperPoint+SuperGlue & {\color{red}166.76 [1.77]} & \color{red}2843.62 \\
& LoFTR & 132.47 [0.19] & 534.0 \\
\midrule

\multirow{6}{*}{
\parbox{1.5cm}{\includegraphics[width=0.6cm]{images/sosmatch_latespring.pdf} 
\includegraphics[trim={8cm 0.5cm 8cm 1cm}, clip, width=0.6cm]{images/summer2.png} \\
$\Delta_T$ = 15 days}
}
& \textbf{Ours} & \cellcolor{yellowcalm!70}8.06 [0.14] & \cellcolor{asparagus!50}{0.05}\\ 
& ORB+RANSAC & 40.38 [0.44] & \cellcolor{yellowcalm!70}25.43\\ 
& SIFT+RANSAC & {64.75 [0.81]} & 407.20\\ 
& Any-Loc & \cellcolor{asparagus!50}{0.12 [0.01]} & 310.05 \\ 
& SuperPoint+SuperGlue & {\color{red}220.87 [4.52]} & \color{red}2907.95 \\
& LoFTR & 154.05 [0.53] & 179.6 \\
\bottomrule

\end{tabular}
\vspace{-5pt}
\label{tab:cross-season}
\end{table}

Our method provides better localization performance than the reference methods in the Summer 1 vs. Summer 2, Early Spring vs. Late Spring, and Late Spring vs. Summer 2 cases. Based on \figref{fig:p-r-curves}, in the Winter 1 vs. Winter 2 cases, SuperPoint+SuperGlue appears to outperform our method. While there is a performance benefit in this case for SuperPoint+SuperGlue, in all cases, our method outperforms all reference methods with a significant margin in terms map size, and it outperforms all reference methods aside from AnyLoc in terms of search time, as seen in \tabref{tab:cross-season}. A low search time is a particularly critical characteristic for the description of a localization approach where search time is critical for the suitability for online implementation. Furthermore, a small map size is a key enabler of collaborative localization, where robots must share their local maps with limited network bandwidth.


\subsection{F1 Score: Performance Analysis by Viewpoint}

\begin{figure}[b!]
    \centering
    \includegraphics[width=\columnwidth, trim={0cm 0cm 13cm 0cm}, clip]{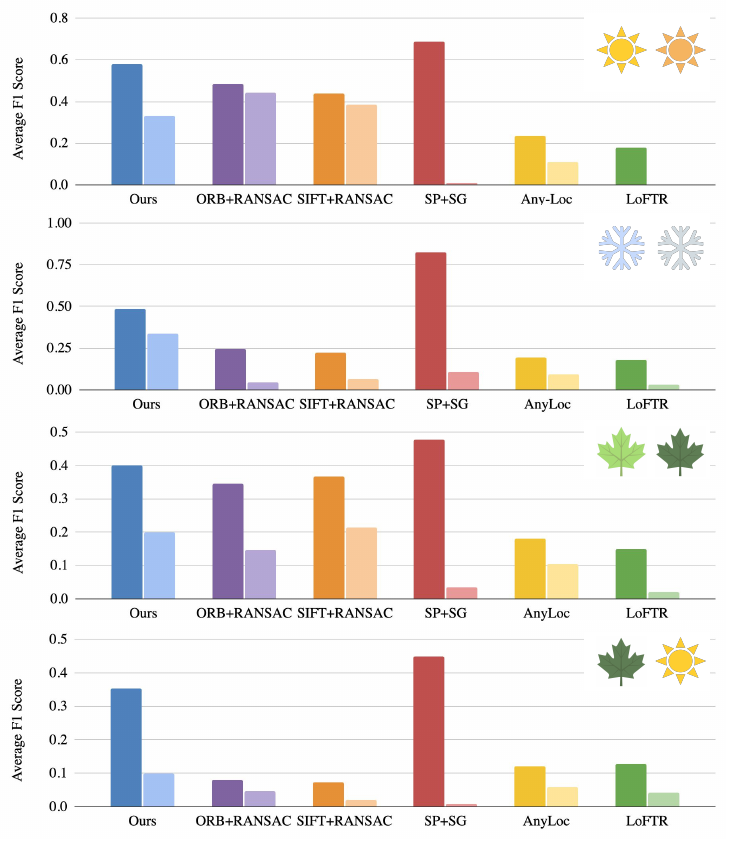}
    \caption{Average F1 scores of different cross-season cases. Bars indicate the performance from the same viewpoint (left) and from different viewpoints (right).}
    \label{fig:f1}
\end{figure}

To further analyze the performance of our method by viewpoint, we consider the cases in which an agent localizes within the map of another agent from the same viewpoint and when an agent passes over a place previously seen by another agent from a separate viewpoint. We calculate the average F1 score of different flights separated into the same viewpoint and different viewpoint, as shown in \figref{fig:f1}.

In this evaluation, we take the average F1 score calculated as the average value between when precision is tuned to at least $0.99$ and when recall is tuned to at least $0.99$. If recall cannot be tuned to at least $0.99$, we start at the highest value. Thus, these results demonstrate performance in cases that benefit from trading off between precision and recall. 

In \figref{fig:f1}, we see that all methods have a lower average F1 score in the different viewpoint setting than in the same viewpoint setting. We note that our method does not perform as well as ORB- and SIFT-based methods in the Summer 1 vs. Summer 2 case from multiple viewpoints due to our map representation being sparse and thus limited by field of view, but it maintains performance while others degrade when there are more visual variations between the flights.

In the same viewpoint case, Superpoint+Superglue surpasses our method in average F1 score. However, in the different viewpoint case, the Superpoint+Superglue fares unfavorably against our method and most comparison methods, suggesting that the approach is very sensitive to variation in viewpoint.

\begin{figure}[b!]
    \centering
    \begin{subfigure}[t]{0.45\linewidth}
        \centering
        \includegraphics[width=1.5in]{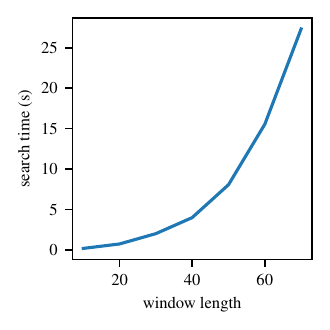}
        \caption{Search time as function of window length.}
    \end{subfigure}%
    ~ 
    \begin{subfigure}[t]{0.45\linewidth}
        \centering
        \includegraphics[width=1.5in]{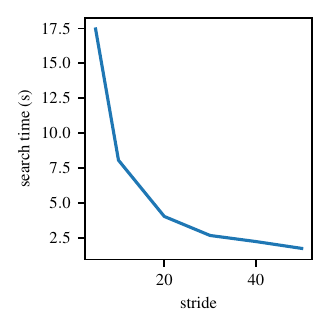}
        \caption{Search time as function of stride.}
    \end{subfigure}
    \caption{Search time evaluations varying window length and stride.}
    \label{fig:searchtimestudty}
\end{figure}

\begin{figure}[b!]
    \centering
    \begin{subfigure}[t]{0.45\linewidth}
        \centering
        \includegraphics[width=1.5in]{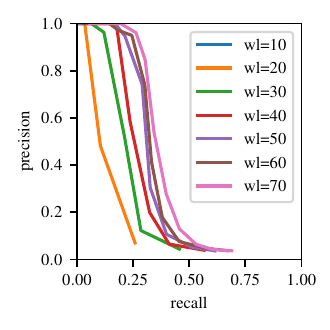}
        \caption{Varying window length, stride $=10$.}
    \end{subfigure}%
    ~ 
    \begin{subfigure}[t]{0.45\linewidth}
        \centering
        \includegraphics[width=1.5in]{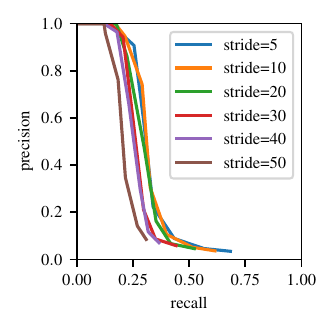}
        \caption{Varying stride, window length $=50$.}
    \end{subfigure}
    \caption{Precision and recall at different values of stride and window length (wl). Each graph is generated by varying minimum count of matches in window.}
    \label{fig:windowstrideprerestudy}
\end{figure}

\subsection{Ablation study}

Due to the computational complexity of the correspondence search problem, we include tunable parameters for the windowed search approach that trade off between performance and runtime. Window length \windowlength{} is defined as the number of objects we consider at a time by their ID as assigned during map construction. Stride length \stride{} is the number of objects we skip as we slide the window along the entire traverse. In \figref{fig:searchtimestudty}, the search time throughout the entire traverse is evaluated. We demonstrate that increasing the window length increases runtime and increasing the stride length, and increasing the stride length decreases runtime.

To evaluate the performance of the system as the window length and stride length are tuned, we look at the precision and recall curves in \figref{fig:windowstrideprerestudy}. Increasing the window length increases the performance of \methodabbreviation{} and decreasing the stride increases the performance of the system.

We find that the performance increases up to a window length of $50$ and a stride length of $10$, reflecting the parameters used in our experiments. These parameters enable tuning as a trade-off between performance and runtime that can be tailored to applications.

\section{Discussion}

\methodabbreviation\ demonstrates the value of incorporating foundation models into front-end object detection and map construction in unstructured environments. Using segmentation in open-set unstructured settings such as dense forested regions provides sufficient geometric cues that are highly suitable for localization and loop closure detection. Our method also offers a significant speed improvement in search time and size reduction in map size in comparison to reference methods. We consider these major improvements towards satisfying the the requirements of a robust description of environment measurements for use in the localization problem, whose requirements we briefly listed in \secref{sec:introduction}.

We share the \batvik{} seasonal dataset, which represents a challenging real-world scenario for visual navigation in unstructured environments, with significant ambiguities in the appearance of the environment. The data includes typical quality issues that occur in drones with hardware constraints such as image compression artifacts, which are useful for real-world evaluation.
Our work reveals that most baseline methods are affected by even short time gaps between traverses, highlighting the need for robust visual approaches in these environments. The release of this dataset enables evaluation of robustness to changing seasons and visual conditions.

Our method does not fully account for uncertainty, and we plan to address cases with less favorable (non-bird's eye view) triangulation geometry that may impact depth accuracy, as well as scenarios with significant odometry drift in future work. 

The comparison of the localization performance against SuperPoint+SuperGlue suggests that it may be possible to find a balance between map size and localization capability by combining the information about the environment's structure with a learning-based approach to description. We thereby plan to incorporate additional information about the objects, including semantic information in environments containing variable discernable objects, incorporating anisotropic object location covariance, and developing robust descriptors for geometry to enable faster correspondence search over large hypothesis spaces. Our proposed method uses size descriptors as a means for excluding putative matches where the size difference of objects is significant, and further search time reduction may be achievable if richer descriptors can be extracted for the segments.

\section{Conclusion}

We present \methodabbreviation, a framework for compact mapping and fast localization that is able to operate in open-set unstructured environments containing segmentable objects. The maps' compactness supports a multi-agent scenario, facilitating efficient communication streams between agents. Experiments with the \batvik{} seasonal dataset demonstrate the pipeline's ability to align frames in a challenging unstructured environment with robustness to temporal and viewpoint variations. \methodabbreviation{} shows that segmentation provides geometric cues suitable for localization and robust correspondence search in unstructured environments.

\section*{Acknowledgement}
This work was supported by Saab Finland Oy, Boeing Research \& Technology, and the NSF GRFP under Grant No. 2141064. The dataset was collected as part of Business Finland project Multico (6575/31/2019). We thank the computational resources provided by the Aalto Science-IT project.

\balance
\bibliographystyle{./bibliography/IEEEtran}
\bibliography{./bibliography/IEEEabrv,./bibliography/bibliography}

\end{document}